\documentclass[conference]{IEEEtran}
\IEEEoverridecommandlockouts
\usepackage{cite}
\usepackage{amsmath,amssymb,amsfonts}
\usepackage{algorithmic}
\usepackage{graphicx}
\usepackage{textcomp}
\usepackage{xcolor}
\usepackage{marvosym}
\usepackage{bm}
\usepackage{bbm}
\usepackage {latexsym}
\usepackage{utfsym}
\usepackage{algorithm,algorithmic,multirow,booktabs}
\usepackage{geometry}
\usepackage{color}
\usepackage{hyperref}

\usepackage{comment}
\def\BibTeX{{\rm B\kern-.05em{\sc i\kern-.025em b}\kern-.08em
    T\kern-.1667em\lower.7ex\hbox{E}\kern-.125emX}}
\begin{document}

\title{SVASTIN: Sparse Video Adversarial Attack via Spatio-Temporal Invertible Neural Networks}

\author{\IEEEauthorblockN{Yi Pan, Jun-Jie Huang, Zihan Chen, Wentao Zhao\textsuperscript{\Letter}, Ziyue Wang}
\IEEEauthorblockA{College of Computer Science and Technology, National University of Defense Technology, Changsha, China\\
\{panyi\_jsjy, jjhuang, chenzihan21, wtzhao, wangzy13\}@nudt.edu.cn}

\thanks{This work is supported by the National Natural Science Foundation of China under Project 62201600 and U1811462, NUDT Innovation Science Foundation 23-ZZCXKXKY-07, and NUDT Research Project ZK22-56.}
}

\maketitle

\begin{abstract}
Robust and imperceptible adversarial video attack is challenging due to the spatial and temporal characteristics of videos. The existing video adversarial attack methods mainly take a gradient-based approach and generate adversarial videos with noticeable perturbations. In this paper, we propose a novel Sparse Adversarial Video Attack via Spatio-Temporal Invertible Neural Networks (SVASTIN) to generate adversarial videos through spatio-temporal feature space information exchanging. It consists of a Guided Target Video Learning (GTVL) module to balance the perturbation budget and optimization speed and a Spatio-Temporal Invertible Neural Network (STIN) module to perform spatio-temporal feature space information exchanging between a source video and the target feature tensor learned by GTVL module. Extensive experiments on UCF-101 and Kinetics-400 demonstrate that our proposed SVASTIN can generate adversarial examples with higher imperceptibility than the state-of-the-art methods with the higher fooling rate. Code is available at \href{https://github.com/Brittany-Chen/SVASTIN}{https://github.com/Brittany-Chen/SVASTIN}.
\end{abstract}


\section{Introduction}
\label{sec:intro}
Recent advances in Deep Neural Networks (DNNs) have enabled effective visual understanding in images and videos~\cite{Karpathy2014LargeScaleVC,Carreira2017QuoVA,Zhang2021VideoLTLL,Zhu2020ACS,Yang2017CatchingTT,Liu2018SibNetSC}.
At the same time, the security issue of DNNs has attracted a lot of attention~\cite{Szegedy2013IntriguingPO,Carlini2016TowardsET,MoosaviDezfooli2015DeepFoolAS,duan2021advdrop,Chen2022ImperceptibleAA, wang2023MPAA, wang2024DeMPAA}. Szegedy~\textit{et al.}~\cite{Szegedy2013IntriguingPO} found that the DNN-based image classifiers can be easily deceived by adding mild adversarial noise to the benign images. Since then different adversarial attack methods~\cite{Szegedy2013IntriguingPO,Carlini2016TowardsET,MoosaviDezfooli2015DeepFoolAS,duan2021advdrop} have been proposed to investigate the security issue of DNNs.  The existing image-domain adversarial attack approaches can be grouped into two main categories: adding adversarial perturbation ~\cite{Szegedy2013IntriguingPO,Carlini2016TowardsET,MoosaviDezfooli2015DeepFoolAS} to the benign images or dropping class specific features from the benign images~\cite{duan2021advdrop}. Recently, Chen~\textit{et al.}~\cite{Chen2022ImperceptibleAA} proposed a Adversarial Attack via Invertible Neural Networks (AdvINN) method to generate imperceptible and robust adversarial examples by utilizing Invertible Neural Networks (INNs) ~\cite{Dinh2014NICENI,huang2021linn,huang2021winnet,Bai2020ExploringAE,Cui2023CertifiedII} to perform feature space feature exchanging and simultaneously drop discriminant information of clean images and add class-specific features of the target images.

\begin{figure*}[t]
    \centerline{\includegraphics[scale=0.48]{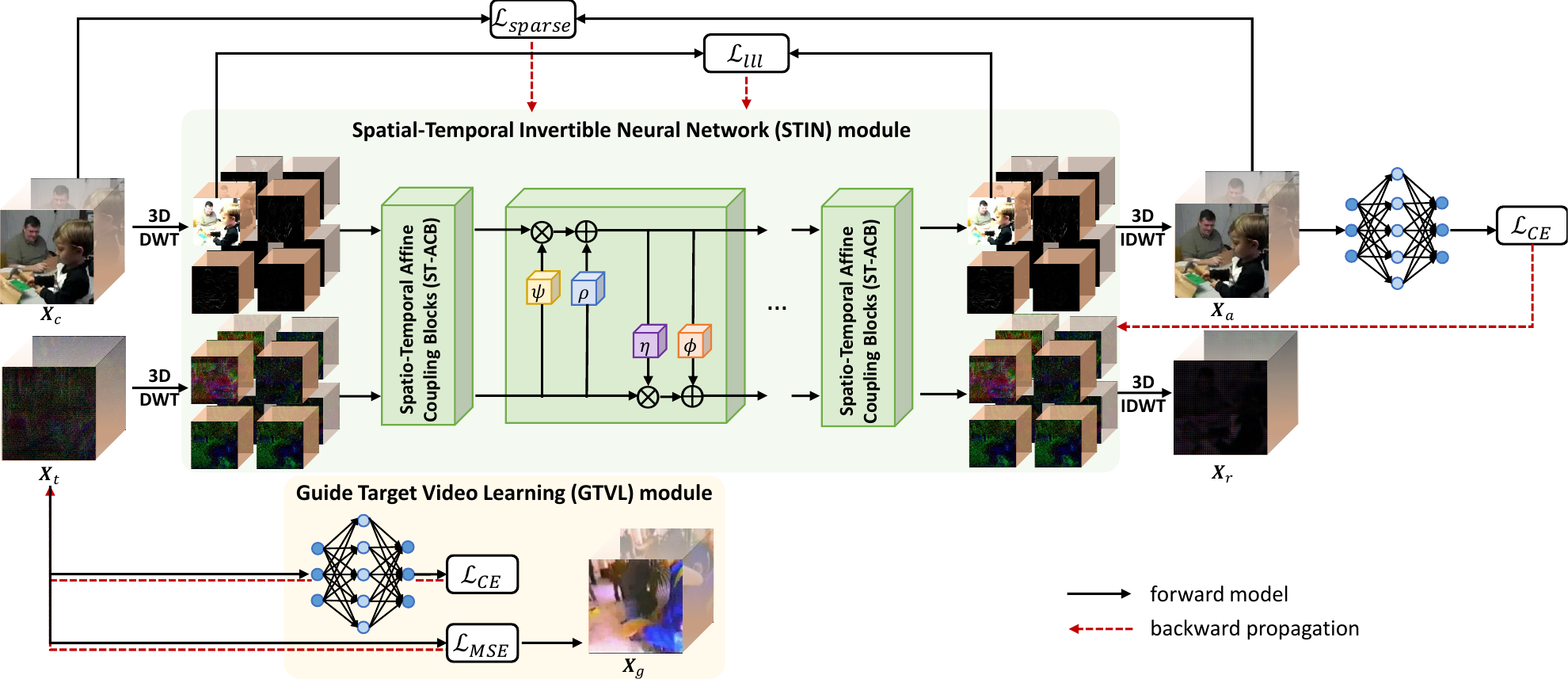}}
    \caption{The overview of Sparse Adversarial Video Attack via Spatio-Temporal Invertible Neural Networks (SVASTIN). The Spatio-Temporal Invertible Neural Network (STIN) module, which utilizes the information preservation property to non-linearly exchange information between the input benign video and the target video. The Guided Target Video Learning (GTVL) module is proposed to update the learnable target video $\bm{X}_{t}$.}
    \label{fig_1}
\end{figure*}
Videos contain both spatial dimensions and a temporal dimension,
and are comprised of a sequence of images with high temporal consistency among adjacent frames. 
The adversarial attack methods for images can be na\"ively extended for videos, however, 
the large searching space and high information redundancy pose a challenge in robustly and effectively generating adversarial video examples.
In order to improve the imperceptibility as well as reduce computational cost, Wei~\textit{et al.}~\cite{wei2019sparse} proposed a sparse attack method for videos by utilizing the $l_{2,1}$-norm to regularize the adversarial noise to be temporally sparse and the generated perturbations are transferred to other frames through the temporal interaction of the target model. 
Inkawhich~\textit{et al.}~\cite{inkawhich2018adversarial} proposed an adversarial attack method for optical-flow-based action recognition based on Fast Gradient Sign Method (FGSM)~\cite{Szegedy2013IntriguingPO}.
To accelerate the speed of adversaries generation for videos, Mu \textit{et al.}~\cite{Mu2021SparseAV} proposed DeepSAVA with an effectively alternative optimization scheme between selecting key frames and optimizing spatial-transformed perturbations.
By utilizing the temporal information, Jiang \textit{et al.}~\cite{Jiang2023TowardsDS} proposed a video attack framework named V-DSA which utilizes optical flow to select the  perturbed pixels of every frame. Although it reduces the search space, the adversarial perturbations are perceptible for human observers.

In order to improve the imperceptibility of adversarial examples, we propose a Sparse Video Adversarial Attack via Spatio-Temporal Invertible Neural Networks (SVASTIN) method which generates adversarial videos through spatio-temporal feature exchanging by utilizing our proposed Spatio-Temporal Invertible Neural Networks.
To accelerate convergence speed, a Guided Target Video Learning (GTVL) module is proposed to learn a discriminative target feature tensor of the target class.
We propose to build a Spatio-Temporal Invertible Neural Network (STIN) module with 3D Discrete Wavelet Transform (3D-DWT) and Spatio-Temporal Affine Coupling Blocks (ST-ACB). It is set to exchange the Spatio-temporal information between the source video and the target feature tensor learned by GTVL module. 
We further constrain the perturbations to be added to the 3D-DWT high-frequency coefficients resulting in adversarial videos with higher imperceptibility.
Extensive experiments on the Kinetics-400~\cite{Kay2017TheKH} and UCF-101~\cite{Soomro2012UCF101AD} datasets demonstrate that the proposed SVASTIN method can generate adversarial videos with the higher fooling rate and higher imperceptibility.

\section{proposed method}
In this section, we introduce the details of Sparse Video Adversarial Attack via Spatio-Temporal Invertible Neural Networks (SVASTIN) which generates imperceptible and robust adversarial videos by fully considering the spatio-temporal characteristics of videos.

\subsection{Overview}
Fig.~\ref{fig_1} illustrates the overview of the proposed SVASTIN method which mainly consists of a Spatio-Temporal Invertible Neural Network (STIN) module and a Guided Target Video Learning (GTVL) module.
In order to better perform spatio-temporal information exchanging, we propose to construct the STIN module with the 3D Discrete Wavelet Transform (3D-DWT) for video decomposition and an Invertible Neural Network with Spatio-Temporal Affine Coupling Blocks (ST-ACB) to capture motion information from both spatial and temporal dimensions. 
It reversibly exchanges information between a clean video $\bm{X}_{c}$ and a target feature tensor $\bm{X}_{t}$, and generates an adversarial video $\bm{X}_{a}$ with a residual video $\bm{X}_{r}$.
The target feature tensor $\bm{X}_{t}$ serves as the source information of the target class.
We propose the GTVL module to effectively learn a target feature tensor $\bm{X}_{t}$ with the guidance of a guide video $\bm{X}_{g}$ and the target classifier $F_{\phi}(\cdot)$. 

\subsection{Spatio-Temporal Invertible Neural Network Module}
The Spatio-Temporal Invertible Neural Network (STIN) module aims to perform spatio-temporal feature space information exchanging between a benign video and a target feature tensor. Invertible Neural Networks (INNs), which is with the information preservation property, has been utilized for imperceptible and robust image adversarial attack~\cite{Chen2022ImperceptibleAA}. Compared to images, videos have an additional temporal dimension which is an essential characteristic and should be exploited. The proposed STIN module is able to extract and process temporal and spatial information simultaneously which consists of a 3D Discrete Wavelet Transform (3D-DWT) layer for video feature decomposition, $M$ ST-ACB for feature updating and prediction, and an inverse 3D-DWT layer for video recomposition.

\noindent  \textbf{3D Discrete Wavelet Transform (3D-DWT) Layer:} Wavelet transform has been widely used for analyzing images and signals at different frequency components.
In this paper, we propose to use 3D-DWT for decomposing both the spatial and the temporal dimensions of a video $\bm{X} \in \mathbbm{R}^{C \times T \times W \times H}$ into low-frequency and high-frequency components 
resulting in 8 sub-bands. Its wavelet coefficients can be represented as $\Gamma \left( \bm{X} \right) \in \mathbbm{R} ^ {8C \times T/2 \times W/2 \times H/2 }$. 
Note that compared with 2D-DWT, 3D-DWT further decomposes the temporal dimension into low-frequency and high-frequency sub-bands, which can help us capture motion information in videos.

\noindent \textbf{Spatio-Temporal Affine Coupling Blocks (ST-ACB):} 
Affine Coupling Block (ACB)~\cite{Dinh2014NICENI} is the key building block of INNs, however, it is designed for processing 2D images and cannot well capture the temporal information in videos. Here, we propose Spatio-Temporal Affine Coupling Blocks (ST-ACB) with 3D convolution layers to promote temporal information modeling.
We denote $\bm{w}^i$ as the input features of the $i$-th Spatio-Temporal Affine Coupling Block, and with $\bm{w}_c^{0} = \Gamma\left( \bm{X}_{c}\right)$ and $\bm{w}_t^{0} = \Gamma\left( \bm{X}_{t}\right)$. Then, the forward process of the $i$-th ST-ACB can be expressed as:
\begin{equation}
\begin{aligned}
    \bm{w}_{c}^{i} =& \bm{w}_{c}^{i-1}  \odot \exp \left(\alpha\left(\psi\left(\bm{w}_{t}^{i-1}\right)\right)\right)+\phi\left(\bm{w}_{t}^{i-1}\right),\\
    \bm{w}_{t}^{i} =& \bm{w}_{t}^{i-1} \odot \exp \left(\alpha\left(\rho\left(\bm{w}_{c}^{i}\right)\right)\right)
+\eta\left(\bm{w}_{c}^{i}\right),
\end{aligned}
\label{eq_4}
\end{equation}
where $\odot$ denotes element-wise multiplication, $\alpha$ is a Sigmoid function multiplied by a constant factor, and $\psi\left( \cdot \right),\phi\left( \cdot \right),\rho\left(\cdot\right),\eta\left(\cdot\right)$ denote dense network architectures extending to 3D convolutional layers~\cite{Wang2018ESRGANES}. 

Given the output of the $M$-th ST-ACB, the adversarial video and the residual video can be reconstructed using inverse 3D-DWT with $\bm{X}_{a} = \Gamma^{-1}\left( \bm{w}_{c}^{M}\right)$ and $ \bm{X}_{r} = \Gamma^{-1}\left( \bm{w}_{t}^{M}\right)$. Therefore, STIN module is fully invertible and the output videos $\left(\bm{X}_{a}, \bm{X}_{r}\right)$ contain the same amount of information as the input videos $\left( \bm{X}_{c}, \bm{X}_{t} \right)$. 

\noindent \textbf{Loss Functions:} 
We use an adversarial loss $\mathcal{L}_a$ to guide the optimization of the parameters of STIN. In detail, $\mathcal{L}_a$ includes a cross-entropy loss for between the predicted class of $\bm{X}_{a}$ and the target class $y_t$, a $\ell_{2,1}$-norm based sparse loss for regulating generated perturbations to be more sparse and a low-frequency loss for constraining target contents concealing to high-frequency coefficients of 3D-DWT which leads to more imperceptible results. It can be expressed as:
\begin{equation}
    \begin{aligned}
            \mathcal{L}_{a} 
        =&\lambda_{a}\ell_{CE}\left( F_{\phi}\left( \bm{X}_{a}\right),y_{t}  \right) +  \frac{\beta_a}{T} \Vert \bm{X}_{a} - \bm{X}_{c} \Vert_{2,1} \\
        &+  \frac{8\gamma_a}{N}\Vert \bm{X}_{c(lll)} - \bm{X}_{a(lll)} \Vert_F^2,
        \label{eq:adv_loss}
    \end{aligned}
\end{equation}
where $N=CTWH$, $\lambda_{a}, \beta_{a}$ and $\gamma_a$ are the regularization parameters, $F_{\phi}\left( \cdot \right)$ is the target classifier, and the subscript ${(lll)}$ denotes the low-low-low-frequency sub-band of 3D-DWT.

\begin{figure*}[t]
    \centering
    \includegraphics[scale=0.62]{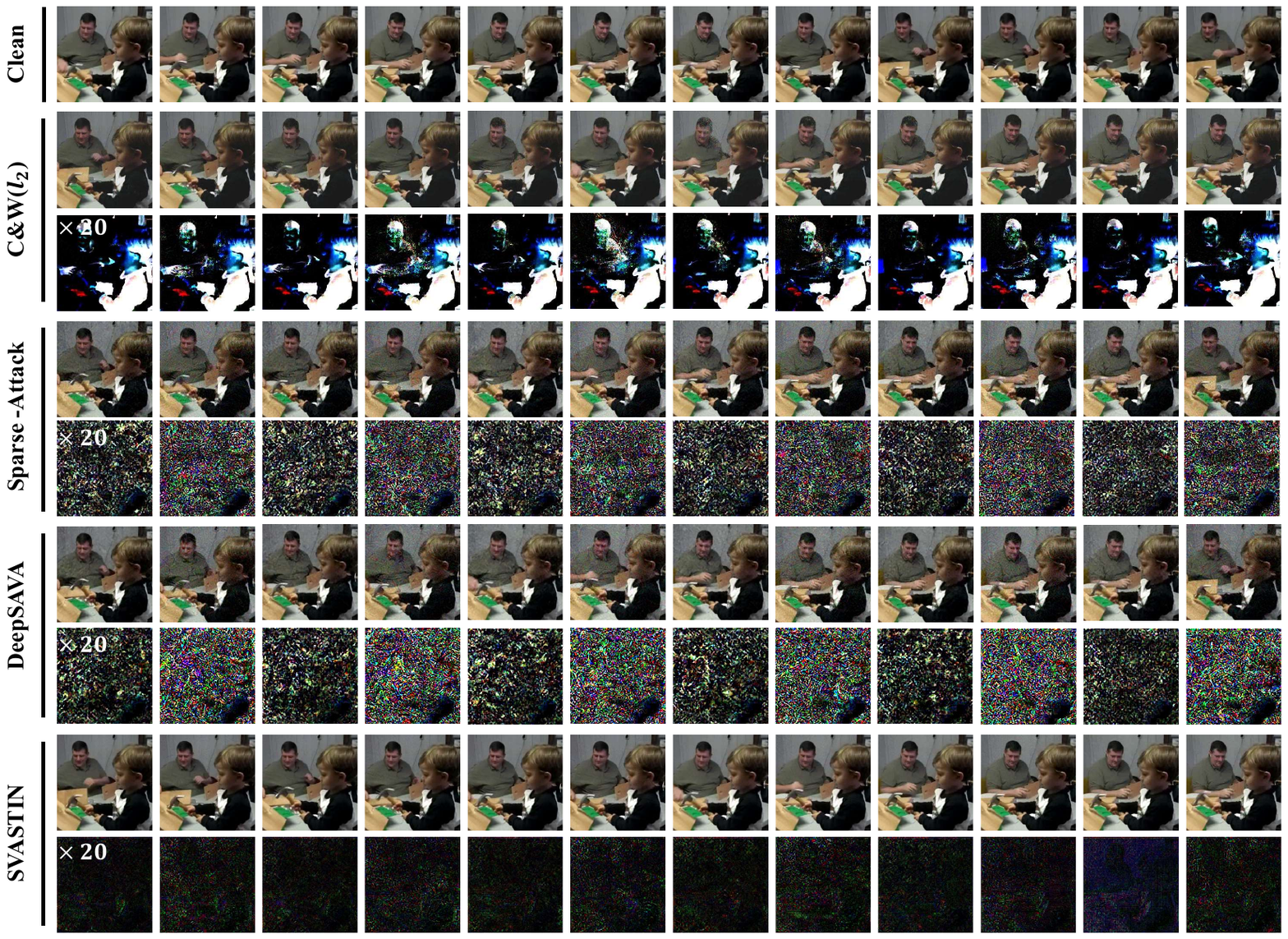}
    \caption{
    Visualization of the generated adversarial video and residual images of different methods
    on UCF-101 dataset against MVIT model. The clean video is successfully classified by the target classifier as \textit{Harming} and the target class is \textit{Shotput}. For each method, the upper row displays the generated adversarial video frames, and lower row shows the residual images which are enlarged by {20} times for better perception.}
    \label{fig_2}
\end{figure*}

\subsection{Guided Target Video Learning Module}
For image adversarial attack, Chen \textit{et al.}~\cite{Chen2022ImperceptibleAA} propose three methods to select target image, \textit{i.e.}, Highest Confidence Target Image (HCT), Universal Adversarial Perturbations Image (UAP) and Classifier Guided Target Image (CGT), however, they cannot lead to effective and efficient target information learning for video adversarial attack.
Specifically, HCT carries a considerable amount of redundant information unrelated to the target class, the UAP approach is unable to successfully generate perturbations for all target classes, and CGT needs to be optimized within a much larger searching space than images.

\noindent \textbf{GTVL Module:}
To improve the effective and efficiency for learning the target video information, we propose a Guided Target Video Learning (GTVL) module to provide suitable target information for STIN module. Specifically, the target feature tensor $\bm{X}_{t}$ is set to be a learnable variable and updated with the guidance of a guide video $\bm{X}_{g}$ of the target class as well as the target classification model $F_{\phi}(\cdot)$.
Compared with directly using the HCT Video as the target feature tensor, GTVL module can alleviate the influence caused by large amount of redundant information within HCT Video. 
Compared with the CGT approach, GTVL module utilizes a guide video and a guidance loss to facilitate the fast convergence.
As a result, the target feature tensor with higher confidence are effectively generated by GTVL module. Detailed experimental results can be found in Section \ref{sec:Experiments}.

\noindent \textbf{Loss Functions:}
In order to effectively embed target contents for information exchanging, we apply a guidance loss $\mathcal{L}_g$ to learn the target feature tensor $\bm{X}_{t}$ which includes a cross entropy loss between the predicted class of $\bm{X}_{t}$ and the target class and a MSE loss $\mathcal{L}_{MSE}$ between $\bm{X}_{t}$ and $\bm{X}_{g}$. 
The guidance loss can be expressed as:
\begin{equation}
    \begin{aligned}
    \mathcal{L}_{g}
    & = \lambda_{b}\ell_{CE}\left( F_{\phi}\left( \bm{X}_{t} \right),y_{t}  \right)+ \frac{\beta_b}{N}\Vert \bm{X}_{t} - \bm{X}_{g} \Vert_F^2,
    \end{aligned}
    \label{eq:upt_loss}
\end{equation}
where $\lambda_{b}$ and $\beta_b$ donates the regularization parameters. 

\section{Experiments}
\label{sec:Experiments}

In this section, we present the quantitative and qualitative results of SVASTIN compared with other state-of-the-art sparse video adversarial methods. Furthermore, we verify the effectiveness of STIN module, GTVL module and the low-frequency loss $\mathcal{L}_{lll}$.

\subsection{Experimental Setup}

\noindent \textbf{Dataset and models.} We evaluate the performance of the comparison methods on Kinetics-400~\cite{Kay2017TheKH} and UCF-101~\cite{Soomro2012UCF101AD}. Following ~\cite{Jiang2019BlackboxAA, Wei2021CrossModalTA, Jiang2023EfficientDB, Wei2021BoostingTT}, we randomly select one video from each category that is correctly classified by the target classifier as the input benign video.  Three commonly used video action recognition models: MVIT~\cite{Li2021MViTv2IM}, SLOWFAST~\cite{Feichtenhofer2018SlowFastNF}, and TSN~\cite{Wang2016TemporalSN}, are used as the target classifier. Specifically, we utilize the pre-trained video action recognition models, including TSN~\cite{Wang2016TemporalSN}, SLOWFAST~\cite{Feichtenhofer2018SlowFastNF} and MVIT~\cite{Li2021MViTv2IM} on Kinetics-400 dataset from mmaction2\footnote{\href{https://github.com/open-mmlab/mmaction2}{https://github.com/open-mmlab/mmaction2}}. Besides, we fine-tune these three models on UCF-101~\cite{Soomro2012UCF101AD} dataset.

\begin{table*}[t]
\small
  \centering
  \renewcommand\arraystretch{1.3}
  \caption{ Comparisons with different methods for targeted attacks on benchmark datasets and models. ($\uparrow$ means the higher the better, and vice versa. The best results are in bold.)}
    \begin{tabular}{l|c|l|rccrrr}
    \toprule
    {Datasets} & {Models} & \multicolumn{1}{c|}{Methods} & {MSE$\downarrow$} & {SSIM$\uparrow$} & {PSNR$\uparrow$} & \multicolumn{1}{c}{$\ell_{2,1}\downarrow$}& FID$\downarrow$ & \multicolumn{1}{c}{FR$\uparrow$} \\
    \midrule
    \midrule
    \multirow{12}[6]{*}{Kinetics-400~\cite{Kay2017TheKH}} 
     & \multirow{4}[2]{*}{MVIT~\cite{Li2021MViTv2IM}} & C\&W$\left( l_2 \right)$~\cite{Carlini2016TowardsET}    & 31.43  & 0.92  & 36.09  & 2.62& 12.84 & 397/400 \\
          &       & Sparse-Attack~\cite{wei2019sparse} & 55.98  & 0.74  & 30.93  & 4.67& 183.45 &
          \textbf{400/400} \\
          &       & DeepSAVA~\cite{Mu2021SparseAV} & 153.66  & 0.66  & 25.80  & 12.83& 76.74 & \textbf{400/400} \\
          &       & SVASTIN  & \textbf{6.93} & \textbf{0.96} & \textbf{40.34} & \textbf{0.58}& \textbf{1.91} & \textbf{400/400} \\
          
    \cline{2-9}          
    & \multirow{4}[2]{*}{SLOWFAST~\cite{Feichtenhofer2018SlowFastNF}} & C\&W$\left( l_2 \right)$~\cite{Carlini2016TowardsET}    & 79.03  & 0.91  & 32.07  & 7.53& 30.65 & 340/400 \\
          &       & Sparse-Attack~\cite{wei2019sparse} & 36.91  & 0.90  & 32.71  & 3.52& 30.80 & 393/400 \\
          &       & DeepSAVA~\cite{Mu2021SparseAV} & 68.71  & 0.87  & 30.30  & 6.59& 44.12 & 395/400 \\
          &       & SVASTIN  & \textbf{7.18} & \textbf{0.96} & \textbf{40.08} & \textbf{0.68}& \textbf{8.10} & \textbf{399/400} \\
          
    \cline{2-9} 
    & \multirow{4}[2]{*}{TSN~\cite{Wang2016TemporalSN}} & C\&W$\left( l_2 \right)$~\cite{Carlini2016TowardsET}    & 35.62  & 0.93  & 34.59  & 2.97& 22.27 & 366/400 \\
          &       & Sparse-Attack~\cite{wei2019sparse} & 21.57  & 0.88  & 35.66  & 1.80& 32.37 & 348/400 \\
          &       & DeepSAVA~\cite{Mu2021SparseAV} & 24.29  & 0.88  & 35.18  & 2.07 & 33.77 & 367/400 \\
          &       & SVASTIN  & \textbf{5.52} & \textbf{0.96}  & \textbf{40.87} & \textbf{0.47}& \textbf{13.16} & \textbf{399/400} \\
          
    \cline{1-9} 
    \multirow{12}[6]{*}{UCF-101~\cite{Soomro2012UCF101AD}} 
    & \multirow{4}[2]{*}{MVIT~\cite{Li2021MViTv2IM}} & C\&W$\left( l_2 \right)$~\cite{Carlini2016TowardsET}    & 557.50  & 0.80  & 27.27  & 46.55& 102.94 & 93/101 \\
          &       & Sparse-Attack~\cite{wei2019sparse} & 57.46  & 0.78  & 31.19  & 4.79 & 80.84 & 91/101 \\
          &       & DeepSAVA~\cite{Mu2021SparseAV} & 24.94  & 0.87  & 34.87 & 2.08& 42.48 & \textbf{101/101} \\
          &       & SVASTIN  & \textbf{8.36} & \textbf{0.95} & \textbf{39.35} & \textbf{0.70}& \textbf{14.36} & \textbf{101/101} \\
    
    \cline{2-9}   
        & \multirow{4}[2]{*}{SLOWFAST~\cite{Feichtenhofer2018SlowFastNF}} & C\&W$\left( l_2 \right)$~\cite{Carlini2016TowardsET}    & 60.06  & 0.82  & 34.21  & 5.73& 140.83 & 45/101 \\
          &       & Sparse-Attack~\cite{wei2019sparse} & 23.82  & 0.90  & 34.90  & 2.27 & 40.59 & \textbf{95/101} \\
          &       & DeepSAVA~\cite{Mu2021SparseAV} & 18.77  & 0.90  & 36.86  & 1.79& 35.11 & 83/101 \\
          &       & SVASTIN  & \textbf{12.79} & \textbf{0.92} & \textbf{37.49}  & \textbf{1.21}& \textbf{34.09} & 91/101 \\
          
    \cline{2-9}      
    & \multirow{4}[2]{*}{TSN~\cite{Wang2016TemporalSN}} & C\&W$\left( l_2 \right)$~\cite{Carlini2016TowardsET} & 96.83  & 0.89  & 30.55  & 8.09 & 34.93 & 99/101 \\
          &       & Sparse-Attack~\cite{wei2019sparse} & 28.87  & 0.86  & 34.23  & 2.41& 62.72  & 92/101 \\
          &       & DeepSAVA~\cite{Mu2021SparseAV} & 67.91  & 0.77  & 30.74  & 5.67& 81.55 & \textbf{101/101} \\
          &       & SVASTIN  & \textbf{9.87} & \textbf{0.94 } & \textbf{38.42} & \textbf{1.02}&\textbf{18.74} & 100/101 \\ 
    \bottomrule
    \end{tabular}%
  \label{Table_2}%
\end{table*}%

\noindent \textbf{Settings.} For STIN, Adam optimizer with learning rate $1e^{-4}$ is used to optimize its parameters with respect to Eqn.\eqref{eq:adv_loss}. We empirically set $\lambda_{a} = 0.3, \beta_a = 0.4$ and $\gamma_{a} = 10$, respectively. And the number of learnable parameters of STIN is 5.11M. As for 3D convolution layers of ST-ACB, the size of the 3D convolution kernel is $\left( 3, 3, 3 \right)$ and the padding is set to $\left( 1, 1, 1 \right)$. For GTVL, Adam optimizer with learning rate $1/255$ is used for optimizing $\bm{X}_t$ with respect to Eqn.\eqref{eq:upt_loss}, and the regularization of parameters $\lambda_b$ and $\beta_b$ is set to $1$ and $0.01$, respectively.

\noindent \textbf{Comparison methods.} 
Three white-box attack methods have been included for comparison, with a traditional adversarial attack C\&W ~\cite{Carlini2016TowardsET} and two sparse attacks methods, including Sparse-Attack~\cite{wei2019sparse} and DeepSAVA~\cite{Mu2021SparseAV}.

\noindent \textbf{Evaluation metrics.} Six metrics are used to evaluate the performance of different methods, including Mean Square Error (MSE), Structural Similarity Index (SSIM) ~\cite{wang2004image}, Peak Signal-to-Noise Ratio (PSNR), Sparsity ($\ell_{2,1}$-norm) ~\cite{wei2019sparse}, Fréchet Inception Distance (FID) ~\cite{heusel2017gans} and Fooling rate (FR). 

\subsection{Evaluation on Targeted Attacks}

Table~\ref{Table_2} shows targeted attack performance of different methods on UCF-101 and Kinetics-400. The generated adversarial examples are considered to be successful if the confidence of the target class is greater than 90\%.
From the results, we can see that the proposed SVASTIN method achieves overall the highest FR while generates adversarial examples with the best  quality compared to other methods, especially on perceptual metrics. 
For example, our proposed SVASTIN method achieves 100\% FR in deception the MVIT model on the Kinetics-400 dataset, while having the higher structural and perceptual similarity to the ground-truth video frames than those of comparison methods. In terms of PSNR, our SVASTIN method is 9.41 and 14.54 higher than the Sparse-Attack and DeepSAVA methods, respectively. 
In terms of the perceptual metric FID, our SVASTIN method is 181.54 and 74.83 lower than the Sparse-Attack and DeepSAVA methods, respectively. 
Moreover, Fig.~\ref{fig_2} shows the visual comparison of our method with three comparison methods. We can see that the proposed SVASTIN method achieves the best visual quality and highest imperceptibility. 

\subsection{Ablation study}
\noindent \textbf{Effectiveness of the STIN module.}
STIN module utilizes both the spatial and temporal information of source and target feature tensor for exchanging feature space information. The performance of video adversarial attack methods will be constrained if the temporal feature of the video is not handled properly. Table~\ref{Table_4} shows the attacking performance of the proposed STIN module with 3D-DWT and ST-ACB and IIEM in AdvINN~\cite{Chen2022ImperceptibleAA} with 2D-DWT and 2D convolutions. From the results, we can show that our proposed STIN module achieves better performances than IIEM in terms of both the quality of the adversarial videos and the optimization speed. 

\begin{table}[t]
\small
  \centering
  \renewcommand\arraystretch{1.3}
  \caption{Ablation study: the effectiveness of STIN module.}
  \scalebox{0.8}{
    \begin{tabular}{cc|rcccccc}
    \toprule
    2D & 3D & \multicolumn{1}{c}{MSE$\downarrow$} & SSIM$\uparrow$ & PSNR$\uparrow$ & $\ell_{2,1}$$\downarrow$ &FID$\downarrow$&Epochs$\downarrow$&FR$\uparrow$\\
    \midrule
    \midrule
    \usym{2714} &    & 18.98  & 0.91  & 35.42  &  1.58 & 5.46 & 55.71&\textbf{400/400}\\
     & \usym{2714}   & \textbf{6.93}  & \textbf{0.96}  & \textbf{40.34}  & \textbf{0.58}  & \textbf{1.91} & \textbf{9.89}&\textbf{400/400}\\
    \bottomrule
    \end{tabular}%
    }
  \label{Table_4}%
\end{table}%

\begin{table}[t]
\small
  \centering
  \renewcommand\arraystretch{1.3}
  \caption{Ablation study: the effectiveness of GTVL module.}
  \scalebox{0.8}{
    \begin{tabular}{c|rccccccc}
    \toprule
    Methods & MSE$\downarrow$ & SSIM$\uparrow$ & PSNR$\uparrow$ & $\ell_{2,1}$$\downarrow$& FID$\downarrow$&Epochs$\downarrow$&FR$\uparrow$\\
    \midrule
    \midrule
    {HCT} &  34.09  & 0.88  & 32.94  & 2.84 & 44.82 &25.60& 162/400\\
     {CGT} &  56.98  & 0.77 & 30.58  & 4.76 & 19.95&11.88&\textbf{400/400}\\
     {GTVL} &  \textbf{6.93}  & \textbf{0.96}  & \textbf{40.34}  & \textbf{0.58}  & \textbf{1.91} &\textbf{9.89}&\textbf{400/400}\\
    \bottomrule
    \end{tabular}%
    }
  \label{Table_5}%
\end{table}%

\begin{table}[t]
\small
  \centering
  \renewcommand\arraystretch{1.3}
  \caption{Ablation study: the effectiveness of the low frequency loss.}
  \scalebox{0.8}{
    \begin{tabular}{c|rccccccc}
    \toprule
    $\mathcal{L}_{lll}$ & MSE$\downarrow$ & SSIM$\uparrow$ & PSNR$\uparrow$ & $\ell_{2,1}$$\downarrow$& FID$\downarrow$&Epochs$\downarrow$&FR$\uparrow$\\
    \midrule
    \midrule
    {\usym{2718}} & 9.19   &  0.95 & 39.28  & 0.77 & 6.85 & \textbf{9.70} &\textbf{400/400} \\
     {\usym{2714}} &  \textbf{6.93}  & \textbf{0.96}  & \textbf{40.34}  & \textbf{0.58}  & \textbf{1.91} &9.89&\textbf{400/400}\\
    \bottomrule
    \end{tabular}%
    }
  \label{Table_7}%
\end{table}%

\noindent \textbf{Effectiveness of the GTVL module.}
Table~\ref{Table_5} shows the performance of three methods for learning target feature tensor. CGT learns the target feature tensor only from the classifer, while HCT utilizes a natural video selecting from target class as the target feature tensor. 
As for GTVL module, the target feature tensor $\bm{X}_{t}$ is initialized with a learnable variable, and then randomly select a guide video $\bm{X}_{g}$ from the target class to further fine-tune $\bm{X}_{g}$ to induce misclassification into target class. We can observe that GTVL module plays an vital role in SVASTIN and effectively improves the imperceptibility of adversarial perturbations and convergence speed.

\noindent \textbf{Effectiveness of the low frequency loss $\mathcal{L}_{lll}$.} 
The 3D-DWT decomposes both the spatial and temporal dimensions of the input videos into 8 low- and high-frequency sub-bands. In order to improve the imperceptibility of adversarial examples, we propose a low-frequency loss to constrain the modification mainly applying to the high-frequency coefficients. It can help to impose smoothness both spatially and temporally.
In Table~\ref{Table_7}, we perform the experiment results on the MVIT model with and without the low-frequency loss $\mathcal{L}_{lll}$. We can observe that SVASTIN with  $\mathcal{L}_{lll}$ demonstrates better qualitative performance in terms of MSE, SSIM, PSNR, $\ell_{2,1}$ and FID, and is with slightly increased optimization speed.

\section{Conclusion}
In this paper, we proposed an adversarial video attack framework, termed as SVASTIN, to generate adversarial video examples based on Spatio-Temporal Invertible Neural Networks (STIN) and a Guided Target Video Learning (GTVL) module. By utilizing the information preservation property of INNs, the proposed STIN Module, driven by the adversarial loss function, performs information exchanging at the spatio-temporal feature level and achieves simultaneously dropping discriminant information of benign video and adding class-specific features of the target feature tensor learned by GTVL to craft adversaries.
Extensive experimental results have shown that the proposed SVASTIN method can generate more imperceptible adversarial examples with higher fooling rate. 

\small
\bibliographystyle{IEEEbib}
\bibliography{icme}

\end{document}